\pdfoutput=1
\documentclass[11pt]{article}

\usepackage[final]{acl}

\usepackage{times}
\usepackage{latexsym}

\usepackage[T1]{fontenc}
\usepackage[utf8]{inputenc}

\usepackage{inconsolata}

\usepackage{graphicx}

\usepackage{algorithm}
\usepackage{algorithmic}
\usepackage[utf8]{inputenc} % allow utf-8 \usepackage[T1]{fontenc}    % use 8-bit T1 fonts
\usepackage{microtype}
\usepackage{graphicx}
\usepackage{subfigure}
\usepackage{booktabs} % for professional tables
\usepackage{subcaption}  % For subfigures
\usepackage{braket} % for formal definition

\usepackage{makecell} % for making cell in the table

\usepackage{multirow}

\usepackage{cuted}

\usepackage{amsmath}
\usepackage{amssymb}
\usepackage{mathtools}
\usepackage{amsthm}

\usepackage[capitalize,noabbrev]{cleveref}
\usepackage[printonlyused]{acronym}
\acrodef{DPO}{Direct Preference Optimization}
\acrodef{HTPS}{HyperTree Proof Search}
\acrodef{GRPO}{Group Reward Preference Optimization}
\acrodef{KTO}{Kahneman-Tversky Optimization}
\acrodef{MCTS}{Monte-Carlo Tree Search}
\acrodef{ORM}{Outcome-supervised Reward Model}
\acrodef{PRM}{Process-supervised Reward Model}
\acrodef{PPO}{Proximal Policy Optimization}
\acrodef{RL}{Reinforcement Learning}
\acrodef{RLHF}{Reinforcement Learning with Human Feedback}

\theoremstyle{plain}

\theoremstyle{definition}

\theoremstyle{remark}

\renewcommand{\algorithmiccomment}[1]{\leavevmode\hfill\makebox[.5\linewidth][r]{//~#1}}

\usepackage{todonotes}

\title{Local Look-Ahead Guidance via Verifier-in-the-Loop for
Automated Theorem Proving}

\author{%
Sara Rajaee\thanks{Equal contribution.}\thanks{
    Work done during an internship at Qualcomm AI Research, Amsterdam.
}\textsuperscript{1},
Kumar Pratik$^*$\textsuperscript{2}, Gabriele Cesa\textsuperscript{2}, Arash Behboodi\textsuperscript{2} \\
Language Technology Lab, University of Amsterdam\textsuperscript{1} \\
Qualcomm AI Research\thanks{
    Qualcomm AI Research is an initiative of Qualcomm Technologies, Inc. \\
    ©2025 Qualcomm Technologies, Inc. and/or its affiliated companies. All Rights Reserved.
}\ , Amsterdam\textsuperscript{2} \\
\texttt{s.rajaee@uva.nl} \\
\texttt{\{kpratik, gcesa, behboodi\}@qti.qualcomm.com} \\
}

\begin{document}
\maketitle
\begin{abstract}
The most promising recent methods for AI reasoning require applying variants of reinforcement learning (RL) either on rolled out trajectories from the LLMs, even for the step-wise rewards, or large quantities of human-annotated trajectory data. The reliance on the rolled-out trajectory renders the compute cost and time prohibitively high. In particular, the correctness of a reasoning trajectory can typically only be judged at its completion, leading to sparse rewards in RL or requiring expensive synthetic data generation in expert iteration-like methods.
In this work, we focus on the Automatic Theorem Proving (ATP) task and propose a novel verifier-in-the-loop design, which, unlike existing approaches that leverage feedback on the entire reasoning trajectory, employs an automated verifier to give intermediate feedback at each step of the reasoning process. Using Lean as the verifier, we empirically show that the step-by-step local verification produces a global improvement in the model's reasoning accuracy and efficiency.
\end{abstract}

\section{Introduction}

As the new applications of modern large language models (LLMs) are emerging in various scientific and engineering domains,  automated mathematical theorem proving has garnered interest from both machine learning researchers and mathematicians. Many ongoing efforts leverage reinforcement learning and expert iteration, inspired by the success of methods like AlphaZero, to build models that search the proof space and provide step-wise or holistic solutions \cite{lample2022hypertree, xin2024deepseek, gloeckle2024abel, anthony2017thinking, doi:10.1126/science.aar6404}.  These solutions are usually verified by formal proof verification systems like Lean \cite{Moura2021-Lean4} or Coq \cite{Coq2024}.  Relying on \ac{RL} is advantageous in terms of data efficiency but comes with high computational and training costs \cite{gloeckle2024abel}. Part of this complexity is related to the necessity of rolling out the proofs and computing rewards from successful episodes. 

In contrast, ReProver \cite{yang2023leandojo} takes a simpler supervised training approach, specifically imitation learning, paired with premise retrieval methods. 
The key components of this proof system are as follows: the theorem we would like to prove, tactics that are actions toward the final proof and itself consist of a set of goals to be proven, premises that are used to prove goals, and the state of the proof which includes the set of goals that are still unproven. 
The approach consists of retrieving the relevant premises from a database given the final theorem and the state of the proof, and then using ReProver to provide tactics for getting to the next state. The proof terminates when all the goals are proven. The method achieves competitive performance with an order of magnitude smaller complexity and training time \cite{gloeckle2024abel}. While the computational cost and simplicity of ReProver are appealing, we empirically observed that many failure cases of tactic generators like ReProver are due to syntactically incorrect tactics or tactics that are not applicable to the current state of a proof. This has detrimental effects on the beam search performed at inference time, as many beams result in invalid tactics that need to be verified by Lean, thus taking away time from exploring more promising tactics.

To preserve the desirable computational efficiency of the ReProver and simultaneously address this problem, the natural choice is to fine-tune the model to remove syntactic errors and increase the number of useful tactics for the proof at each state. Recently, feedback-based alignment has been gaining attraction in various other similar fields, such as automated code generation and various preference optimization methods where the rewards come either from human feedback (RLHF) or other reward models \cite{Ouyang2022-rlhf, Ziegler2019-human-preference,rafailov2024direct}.  Since applying many such reinforcement learning methods for training large models can be complex and expensive, various approaches have been introduced in the literature with moderate complexity, among which we can refer to \ac{DPO}~\cite{rafailov2024direct} and \ac{GRPO}~\cite{Shao2024-deepseekmath}. In the context of mathematical reasoning and theorem proving, many works have emphasized the importance of trajectory-level preferences in mathematical problem solving with large language models (LLMs) (e.g., see \cite{Xiong2024-mulit-DPO} and Preference Optimization paragraph in Sec.~\ref{sec:relatedworks}). Even when complexity is saved in the \ac{RL} training algorithms, computing these trajectory-level preferences can incur additional complexity. This discussion extends to more general episodic reasoning tasks with stepwise verification, where the model needs to provide outputs that are both syntactically and semantically correct and useful for solving the problem at hand. 

\paragraph{Our contribution.} In this work, we aim to address the above issues by fine-tuning a pre-trained prover, which listens and uses the feedback from the tool, in this case Lean, during the training. At each step, our framework, called LeanListener, obtains feedback on the generated tactics directly via its interaction with the Lean software, and performs policy optimization with a reward that is designed based on Lean feedback. Given the pre-trained tactic generator from \cite{yang2023leandojo}, we sample different tactics from its output for each proof state in the training set and use Lean feedback to compute the reward. The reward consists of a negative return for invalid tactics, a positive one for applicable tactics, and a return based on the number of remaining unsolved goals. The \ac{RL} training is done using \ac{GRPO}. First, note that the sampling step for the generator’s output can yield applicable and new tactics that differ from the provided human label. Therefore, it helps the data efficiency of the method by exploring and adding new tactics like what we see in expert iteration. Second, unlike methods like \ac{PRM}, we do not compute the step-wise reward based on the full trajectory information and only rely on \textit{local look-ahead} feedback from the number of remaining and unsolved goals in the next immediate steps and, therefore, we address the complexity of the trajectory based preference association. Thanks to this fine-tuning strategy, we expect the model to rank valid and effective tactics higher than invalid ones, even if they were not previously observed in the human-labeled trajectories. As a result, our fine-tuned model can make better use of the beam search used at inference time. 
Besides, as we will show in our numerical results, the local look ahead, online training with Lean-in-loop, and GRPO are crucial components in improving the final performance of the model. To summarize, our contributions are as follows. We propose a framework for efficient training of a tactic generator LLM to leverage the feedback provided by the external tool, in this case, the Lean solver. We use \ac{GRPO} for training the model using the reward that is based on the applicable tactics and the number of unsolved goals.  The \ac{GRPO} training is particularly beneficial compared to \ac{DPO} training in our case. We show that the training based on offline dataset generation of positive-negative samples is ineffective, and online training is crucial. The online verification from the tool automatically adds new tactics to the training data. Finally, we can obtain noticeable gains by computing the reward using local look-ahead without recourse to the trajectory level preference.

\begin{figure*}[th]
    \centering
    \includegraphics[width=\linewidth]{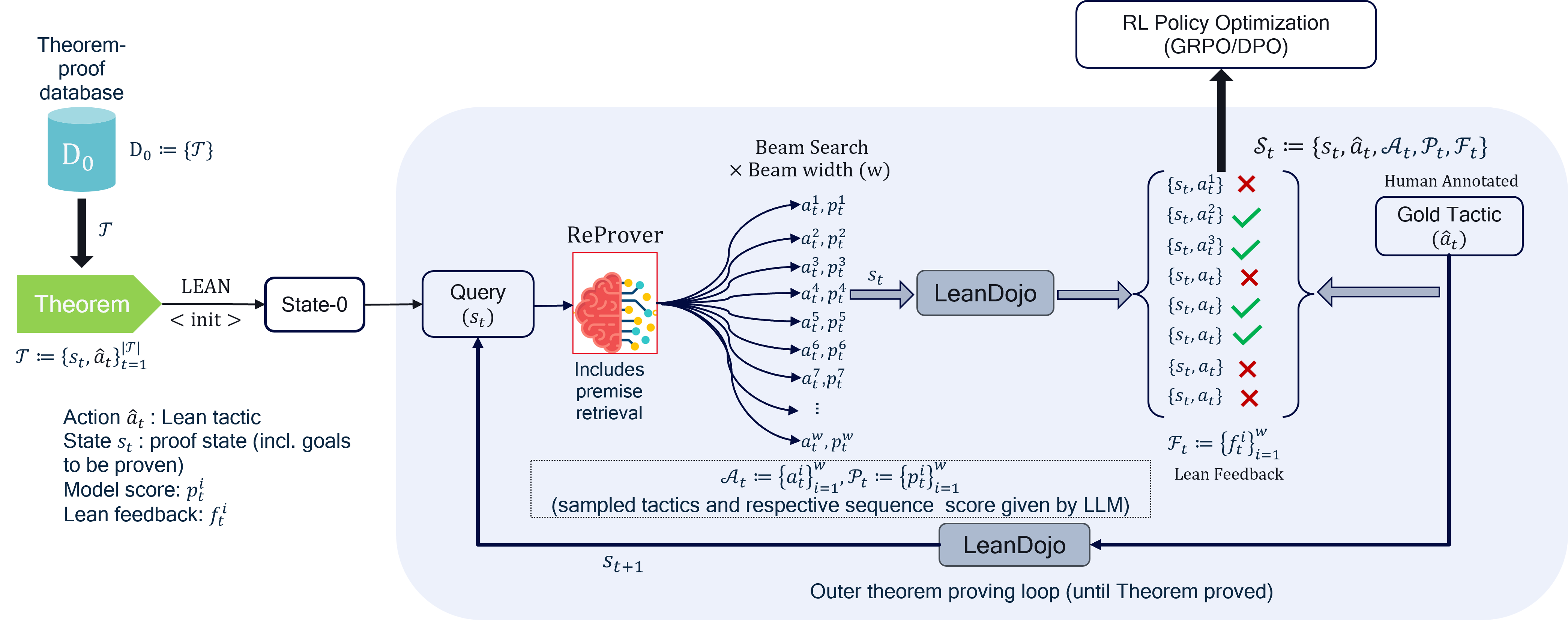}
    \caption{We start with a dataset of human-annotated theorems and their proofs, including the gold tactics, in Lean. During training, the ReProver retrieves the relevant premises from the premise database, and based on those and the state of the proof, generates the next tactics. The generated tactics are verified by Lean, and its feedback is used by RL training loop to fine-tune the ReProver. The loop continues using the steps from the human-annotated tactics until proof termination.}
    \label{fig:data_generation}
\end{figure*}

\section{Related Works}
\label{sec:relatedworks}

In this section, we discuss the closest works related to our paper. We included an extended version of the related works in App. \ref{supp:related_works}, and in particular summarize some works in Table \ref{tab:comparison_provers}.

\paragraph{Theorem Proving.} Large Language Models (LLMs) are currently used in automatic theorem proving to generate the whole proof, or provide assistance in sequential proofs by finding the right premise or right tactics, or assisting in theorem prover specific formalization of mathematical statements, as well as combination of these ideas; see  \cite{xin2024deepseek,Xin2024-deepseekproverv15, Polu2020-gptf, Jiang2022-Thor, Wang2023-LEGO,Jiang2022-DraftSketch,Wu2022-Autoformalization} for some pointers. 
A key component for theorem proving is the formal proof management systems and verifiers like Lean \cite{Moura2021-Lean4} and Coq \cite{Coq2024}. 
% Lean has received particular attention in the mathematics community for example by its use in verification of some components in the condensed mathematics research program \cite{Scholze2022-qj}. 

% We use Lean in this work, and a framework that consists of a premise retrieval part and a prover that generates step-wise proof tactics. In our paper, we are focused on the logic behind tactic generation. In this sense, our approach can be complementary to any work on retrieval and search steps. 

% Different transformer-based models can be used to find useful premises, definitions, and theorems \cite{Irving2016-DeepMath}. 
In Thor \cite{Jiang2022-Thor}, the authors introduce a class of methods to use automated theorem provers for premise selection. LeanDojo \cite{yang2023leandojo} provided a framework that retrieves the premise from a database of Lean premises and uses ReProver, an encoder-decoder model, to generate tactics for theorem proving in Lean. 
LeanAgent \cite{kumarappan2024leanagent} adds a dynamic database to LeanDojo, which enables the continual learning of the agent. Besides, a new curriculum learning based on the difficulty of the theorems is also added. 
GPT-\emph{f} was introduced in \cite{Polu2020-gptf} using Metamath as the formalization framework. Besides, they showed that iterative training a value function on proofs generated by the LLM prover can continually improve the performance.
Another example is LLEMMA, which is based on pre-trained Llama Code \cite{azerbayev2024llemma}. 
DeepSeek-Prover \cite{xin2024deepseek} generated Lean4 proof data by translating natural language problems into their formal versions in Lean4, and the model produces whole proofs in single turns.
DeepSeekV1.5 \cite{Xin2024-deepseekproverv15} provides a middle ground by generating the whole proof, verifying it via Lean, and then truncating the proof until the first error. The authors propose additional techniques relying on proper appending of previous states and including truncate-and-resume in the \ac{MCTS} procedure.
Additionally, DeepSeekV1.5 leverages verification feedback from Lean on whole-proofs to improve model's \emph{alignment with the formal structure} of Lean.
In particular, it uses the Group-Relative Policy Optimization (GRPO) algorithm originally introduced for DeepSeek-Math in \cite{Shao2024-deepseekmath}, which removes the need for a separate critic model, thereby simplifying the reinforcement learning pipeline and reducing its computational cost.
In comparison, we consider a \emph{step-wise} approach, leveraging the verifier feedback at each intermediate step of~the~proof.

\paragraph{Preference Optimization.}
% Another important step in theorem proving, and many other alignment related tasks, is to \textit{align} the output of a given model based on the positive-negative preference pair of samples. 
Other works in the literature have considered the idea of model alignment to improve the reasoning capabilities of language models, too.
In general, there are online versions of \ac{RLHF} as in \cite{Bai2022-rlhf,Ouyang2022-rlhf}, and offline versions with either an explicit reward model (e.g. \cite{Christiano2017-human-preference,Ziegler2019-human-preference}) or an implicit reward model (e.g. \ac{DPO} \cite{rafailov2024direct}) to encode the preference.
The later method directly optimizes the model without training any independent reward model.
% In particular \ac{DPO} uses Bradley-Terry preference model \cite{Bradley1952-bt} and the fact that the optimal solution to the KL-constrained reward maximization objective is known in closed form to simplify the training loss without dependence on a reward model \cite{rafailov2024direct}.
Many follow-up works explored \ac{DPO} variations and its shortcomings, such as reward over-optimization, for example, by introducing \ac{KTO} \cite{Ethayarajh2024-kto}.
% and introducing trust-region based versions of previous alignment models \cite{Gorbatovski2024-ref-model}. 

% Generally, direct preference learning algorithms have recently proven particularly successful to fine-tune and align large open-source models, e.g. \cite{tunstall2024zephyr}.
% It seems to be beneficial to use on-policy sampling and online exploration for direct preference learning

%Different forms of direct preference optimization have been explored in the recent literature for enhancing the reasoning capabilities of language models, in particular for mathematical problem solving,
In the context of reasoning and, in particular, mathematical problem solving, many recent works have explored different forms of direct preference optimization, for example \cite{Xiong2024-mulit-DPO,yuan2024advancing,jiao2024learning,Cobbe2021-train-verifier,Lightman2023-s2sverify,Wang2024-math-shepherd,pang2024iterative,chen2024step,lu2024step}. The authors in \cite{Xiong2024-mulit-DPO} introduce a multi-turn version of \ac{DPO} to use feedbacks from the verifiers, in their case the code interpreters, particularly for multi-turn scenarios, which requires trajectory-level preference optimization. The trajectory-level preference can be obtained by dataset labels of the gold answers or \acp{ORM} \cite{Cobbe2021-train-verifier,Lightman2023-s2sverify}.  A more fine-grained step-wise supervision can be used as in \acp{PRM} \cite{Lightman2023-s2sverify} or  by leveraging trajectory level preferences \cite{Wang2024-math-shepherd}. The idea of \ac{ORM} and \ac{PRM} has also been originally discussed in \cite{Uesato2022-prm-orm}. These works focus on mathematical reasoning, rather then formal theorem proving, which is the focus of our work. In our work, the theorem prover automatically provides the preference, which is used either during training or for an offline generated dataset. Furthermore, our work follows \ac{PRM} philosophy as we consider feedback from Lean at each step of proof generation.

\section{Methodology}
In this section, we begin with an overview of Interactive Theorem Proving (ITP), outlining the task formulation and key terminology. We then introduce our framework, \textit{LeanListener}, and describe its main components.
\subsection{Background}
% \paragraph{Lean Environment}
In ITP, an LLM-based prover interacts with an external proof assistant and receives feedback on the steps to be taken to prove the given theorem.
In this work, we use the proof assistant software \textit{Lean}\footnote{We use Lean 4, and for brevity, simply refer to it as Lean in the rest of the manuscript.} \cite{Moura2015TheLT,Moura2021-Lean4} as the formal environment which LLMs employ to verify each proof step in a proof sequence.
In particular, we used the open-source LeanDojo framework \cite{yang2023leandojo}, which provides toolkits to interact with Lean, as well as readily extracted data and pre-trained models.

An agent can interact with Lean to prove a theorem $T$ by iteratively observing the current "proof state" $s_t$ and performing an action $a_t$, called a "tactic" (i.e., proof steps), which leads to the next state $s_{t+1}$.
Lean inspects each tactic, accepts or rejects it, and provides additional information, like the number of goals yet to be proven. 
Our model can utilize the feedback received from Lean during the training and for the next tactic generation.
Note that Lean is a formal language; hence, it has a strict syntax, and only certain tactics can be applied on each state.
At each step, the agent relies only on the Lean feedback about the current state to generate the next tactic until the theorem $T$ is proved or the model runs out of predefined time. Figure \ref{fig:lean_example} presents a Lean proof example.
Finally, the sequence of state-tactic pairs $\{\langle s_t, a_t \rangle\}_t$ composes the proof of the original theorem $T$.

{
In general, a certain state $s_t$ can include multiple (sub-) goals, representing a number of independent statements to be simultaneously proved in order to prove the original statement.
An intuitive example is given by application of an \emph{induction} tactic, which turns a single statement into two independent ones (the base case and the inductive step).
Hence, the application of a tactic $a_t$ on a certain state $s_t$ can prove certain goals but also turn some into more sub-goals.
As a result, the number of sub-goals $\mathcal{G}(s_{t+1})$ can increase, decrease or remain unchanged after the application of each tactic.
}

% \paragraph{model architecture, task definition, related work for SF}
Tactic prediction is one of the primary tasks in training LLM provers \cite{yang2023leandojo,welleck2023llmstep,lample2022hypertree}.
In a supervised learning framework, during training, the model is fed the current state ($s_t$) and asked to predict a tactic ($a_t$). Recent studies have shown that providing related premises, in addition to the current state, can also enhance the model's performance \cite{mikula2023magnushammer,yang2023leandojo}.
In the inference phase, however, for each theorem, the model employs an inference time-compute technique. It observes the current state ($s_t$) and generates a set of tactics ($\set{a_0,...,a_k}$) using \textit{beam search} with a size of $k$. Building a proof search tree, the tactic generator interacts with the Lean assistant, starting with the tactic with the highest accumulative log probability (best first search). The returned state can be an error state if the tactic execution is unsuccessful, e.g., due to timeout or inapplicable tactic. In this case, the model explores the next tactic in the queue. This process will continue until we reach a valid next state ($s_{t+1}$) or prove the given theorem. It is worth mentioning that the generated proof by the model can be different from the human-written one in the test dataset. 
As many inference-time compute approaches, the model (generator) utilizes feedback from an external expert (verifier) and can explore creative proofs at the cost of extra computation at inference.

\looseness=-1
While the pre-training objective is a supervised sequence-to-sequence task, in the evaluation phase, the model 
acts like an RL agent, which receives feedback on its actions (generated tactics) and uses it to choose the next action via the search strategy explained above.

In this work, we resolve this discrepancy and bring the proof-tree expansion-based exploration to the training regime, leveraging the feedback provided to each tactic to construct step-wise rewards.
Inspired by human preference in LLMs alignment, we present a novel framework, LeanListener, in which we align the pre-trained tactic generator with guidance from Lean in a step-wise scenario. In contrast with similar works with RL training like \cite{Xin2024-deepseekproverv15}, we only need a single step look ahead and do not utilize the full trajectory to compute the rewards.
% \todo{Other works do RL for fine-tuning Theorem-Proving models, we are not the first doing this. Should emphasise more on the step-wise feedback}
In what follows, we explain our LeanListener framework in detail.

\subsection{LeanListener}
\label{sec:methodology}
Our LeanListener framework is based on employing external proof assistant feedback as a guidance signal. 
{
To bridge the discrepancy between the seq2seq-based training and beam search-based proof full tree expansion during inference, we bring the per-step proof-tree expansion based exploration to the training regime.}
Aside from transitioning to the next proof state, applying a tactic in Lean results in valuable information, which can be used to score the efficacy of the tactic at the concerned proof step, and consequently guide the model toward proving theorems. 
For example, a syntactically valid tactic can be gauged for the number of sub-goals in the theorem it helps resolve.

Lean feedback includes helpful information that can be employed to guide the model toward proving theorems, such as if the generated tactic is applicable to the current state and how many (sub-)goals are solved by applying the tactic.
Then, LeanListener utilizes this to reward the model not only for generating an applicable and syntactically correct tactic but also for encouraging the model to generate tactics that solve more (sub)goals.
This results in a more efficient proof-generation process.

A straightforward approach to do so is employing \acf{DPO} \cite{rafailov2024direct}, a popular lightweight but effective algorithm used to align LLMs with human preferences, which is often replacing reinforcement learning based solutions with separate reward models like \ac{RLHF}.
The key ingredient in DPO is creating a static preference dataset in the form of pairwise comparisons $D=\{(x_i, y_i^+, y_i^-)\}_i$ such that for each \emph{prompt} $x_i$, the generated \emph{output} $y_i^+$ is preferred over $y_i^-$. DPO directly optimizes the policy $\pi_\theta$ (possibly, initialized to a reference policy $\pi_{ref}$) via the following loss based on the static dataset:

\begin{small}
\begin{align}
\label{eq:dpo_loss}
    & L_{DPO}(\pi_\theta, \pi_{ref}) = 
    - \mathbb{E}_{(x, y^+, y^-) \sim D} \\
    & \left[
        \log \sigma \left( 
                        \beta \log\frac{\pi_\theta(y^+|x)}{\pi_{ref}(y^+|x)} 
                      - \beta \log\frac{\pi_\theta(y^-|x)}{\pi_{ref}(y^-|x)}
                    \right)
    \right]  \nonumber
\end{align}
    
\end{small}
where $\sigma$ is the logistic function. The DPO preference dataset can be built based on applicable/not applicable tactics as the positive and negative samples in a pair. In essence, the model can be trained to prefer the applicable tactics, i.e., tactics that lead to a valid proof state, over inapplicable tactics, i.e., tactics that lead to an error state. The unsophisticated DPO pairing configuration will intrinsically reward all the applicable tactics equally, irrespective of whether it takes the proof toward a conclusive state or not.
However, a more intricate scoring of tactics can be performed by looking deeper into the Lean feedback, such as the number $\mathcal{G}(s_{t+1})$ of theorem sub-goals that remains to be proven at step $s_{t+1}$. 
To craft a more differential scoring of applicable tactics, as reward $r_t$ for a tactic $a_t$, we choose $R(a_{t}; s_{t}) = \textit{softplus}\left(\mathcal{G}(s_{t}) - \mathcal{G}(s_{t+1})\right)$, where $\textit{softplus}(x, \beta) = \frac{1}{\beta} \ln(1 + e^{\beta x})$. We invariably score the invalid tactics with a score of $0$. Such a scoring strategy rewards, and hence, trains the model to prioritize generating tactics that help resolve more sub-goals. We choose \textit{softplus} over a naive linear delta to prevent negative reward values, since the number of intermediate sub-goals may temporarily increase before the proof is completed.

To take the most out of the Lean feedback and incorporate the number of proven (sub-) goals by the tactics as well, we need a more sophisticated objective. In this regard, we propose to use \emph{Group Relative Policy Optimization} (GRPO) \cite{Shao2024-deepseekmath}, a variant of reinforcement learning (RL) Proximal Policy Optimization (PPO) algorithm \cite{Schulman2017ProximalPO}, in which each tactic is rewarded based on the sub-goals criterion discussed above. For each prompt $q$ (current state $s_t$), GRPO samples a group of outputs $\{o_1, o_2, \cdots, o_G\}$ (generated tactics) from the old policy $\pi_\theta{}_\text{old}$ and maximizes the policy model with the following objective:

\begin{small}
\begin{align}
\label{eq:grpo_loss}
J_{\text{GRPO}}(\theta) = & \; \mathbb{E}_{
                                % \bigg[
                                q \sim P(Q), \{o_i\}_{i=1}^G \sim \pi_\theta{}_\text{old}(O|q)
                                % \bigg]
                                } 
                           \frac{1}{G} \sum_{i=1}^G \frac{1}{|o_i|} \sum_{t=1}^{|o_i|} \bigg\{\\
                          &\min \bigg[ \frac{\pi_\theta(o_{i,t}|q, o_{i,<t})}{\pi_\theta{}_{\text{old}}(o_{i,t}|q, o_{i,<t})} \hat{A}_{i,t}, \nonumber \\
                          & \; \text{clip} \bigg( \frac{\pi_\theta(o_{i,t}|q, o_{i,<t})}{\pi_\theta^\text{old}(o_{i,t}|q, o_{i,<t})}, 1-\epsilon, 1+\epsilon \bigg) \hat{A}_{i,t} \bigg] \nonumber \\
                          & - \beta D_\text{KL}[\pi_\theta || \pi_\text{ref}] \bigg\} \nonumber
\end{align}
    
\end{small}
Where $\pi_\theta$ and $\pi_\theta{}_\text{old}$ are current and old policy models, and $q, o$ are questions and outputs sampled from the dataset and the old policy $\pi_\theta{}_\text{old}$, respectively. $\epsilon$ and $\beta$ are hyper-parameters, and $\hat{A}_{i,t}$ is the advantage calculated based on the reward for each sampled tactic. 
More specifically, assume the reward value for all generated tactics in a beam search with a size of $w$ is represented by $r = \{r_1, r_2, \cdots, r_w\}$. Then, the advantage $\hat{A}_{i,t}$ is the normalized reward $\hat{A}_{i,t} = \tilde{r_i} = \frac{r_i - \text{mean}(r)}{\text{std}(r)}$. Following \cite{Shao2024-deepseekmath}, we estimate the KL divergence with the following equation: 

\begin{small}
\begin{align}
    D_\text{KL}&(\pi_\theta \, || \, \pi_\text{ref}) = \\
    & \frac{\pi_\text{ref}(o_{i,t} | q, o_{i, <t})}{\pi_\theta(o_{i,t} | q, o_{i, <t})} 
    -  \log\left(\frac{\pi_\text{ref}(o_{i,t} | q, o_{i, <t})}{\pi_\theta(o_{i,t} | q, o_{i, <t})}\right) - 1 
    \nonumber
\end{align}
\end{small}

We use the pre-trained generator in the ReProver model from \cite{yang2023leandojo} both as a base reference model for $\pi_{ref}$ and to initialize the policy to optimize $\pi_\theta$.

\begin{table*}[ht]

    \centering
    \scalebox{0.9}{
    \begin{tabular}{lc|cccccc}
         \textbf{Strategy}        &  dropout   &  Prec.$@8$ $\uparrow$     &    MAP          &       MRR       & \makecell{Len. Valid \\ Tactics}    & \makecell{Len. All\\ Tactics}    &  \% 0-precision steps $\downarrow$ \\  
         \hline
         % Offline DPO (rand)       &  $p=0.3$   &  $0.x$          &  $0.x$          &  $0.x$          &   $0.x$               &  $0.x$              &  $x$          \\ 
        \makecell{Offline DPO\\ (zero acc.) } &  $p=0.3$   &  $37.75$       &  $60.75$       &  $64.43$       &   $26.66$           &  $111.20$           &  $29.10$          \\ 
        \makecell{ Offline DPO\\ (hard)}       &  $p=0.3$   &  $35.71$       &  $59.90$       &  $63.42$       &   $28.59$           &  $116.00$           &  $29.80$          \\ 
         % Offline DPO (rand)       &  /         &  $0.x$          &  $0.x$          &  $0.x$          &   $0.x$               &  $0.x$              &  $x$          \\ 
         % Offline DPO (zero acc.)  &  /         &  $0.x$          &  $0.x$          &  $0.x$          &   $0.x$               &  $0.x$              &  $x$          \\ 
        \makecell{ Offline DPO \\ (hard)}       &  /         &  $34.86$       &  $57.92$       &  $61.08$       &   $29.37$           &  $115.39$           &  $32.14$          \\ 
         \hline
         % Online  DPO (rand)       &  $p=0.3$   &  $0.x$          &  $0.x$          &  $0.x$          &   $0.x$               &  $0.x$              &  $x$          \\ 
         % Online  DPO (zero acc.)  &  $p=0.3$   &  $0.x$          &  $0.x$          &  $0.x$          &   $0.x$               &  $0.x$              &  $x$          \\ 
         \makecell{Online  DPO \\ (hard)}      &  $p=0.3$   &  $44.75$       &  $64.99$       &  $71.22$       &   $19.72$            &  $27.44$           &  $11.88$          \\ 
         \hline
         % Online  GRPO             &  $p=0.25$   &  $\textbf{50.225}   $       &  $\textbf{70.26}$       &  $\textbf{77.12}   $       &   $\textbf{16.14}   $            &  $\textbf{19.79}   $           &  $\textbf{7.2}   $          \\ 
         % Online  GRPO             &  $p=0.25$   &  $\textbf{50.47}   $       &  $\textbf{70.16}$       &  $\textbf{77.08}   $       &   $\textbf{16.05}   $            &  $\textbf{19.98}   $           &  $\textbf{7.67}   $          \\ 
         % Online  GRPO             &  $p=0.25$   &  $\textbf{51.43}   $       &  $\textbf{70.81}$       &  $\textbf{77.69}   $       &   $\textbf{15.92}   $            &  $\textbf{19.74}   $           &  $\textbf{7.24}   $          \\ 
         Online  GRPO             &  $p=0.25$   &  $\textbf{51.01}   $       &  $\textbf{70.09}$       &  $\textbf{77.05}   $       &   $\textbf{16.26}   $            &  $\textbf{19.87}   $           &  $\textbf{7.44}   $          \\ 
         \hline
         \makecell{ReProver \\ (base model)}    &  /         &  $40.77$       &  $59.85$       &  $65.70$       &   $18.08$           &  $22.34$            &  $12.74$          \\  
         \hline
    \end{tabular}
    }
    \caption{The DPO objective optimizes the model to prefer valid tactics. In this table, we report different metrics to measure the step-wise performance. Prec.$@8$ = the percentage of the number of valid tactics among the top $8$ tactics sampled at each proof state. MAP= percentage of mean average precision. MRR = percentage of mean reciprocal rank. Len. Valid Tactics = average length (in terms of tokens) of the generated tactics which are valid. Len. All Tactics = average length of all generated tactics. \%$0$-precision steps = fraction of states with no valid tactics.}
    \label{tab:stepwise_stats} 
\end{table*}

\subsubsection{Dataset Curation}
For both DPO and GRPO finetuning, we build our training dataset using the training split of LeanDojo Benchmark \cite{yang2023leandojo} to avoid data contamination as the ReProver model has already been pre-trained on it. Inspired by the online-RL training paradigms for eliciting reasoning in LLMs \cite{gloeckle2024abel}, we, too, build our training paradigm in an online setting.
During online training, we update the model used in the beam search to generate tactic proposals for every $50$ training iteration with the target model. We also experiment with the offline setting, where we use the pre-trained ReProver model to generate the training data offline. We discuss the relative advantages of online over offline training in Table~\ref{tab:stepwise_stats} and Sec.~\ref{sec:experiments}.

\looseness=-1
To generate the fine-tuning data, we parse the human-annotated proofs in the LeanDojo benchmark using either the reference ReProver model (offline) or the target model (online). More specifically, for each proof state $s_t$ in human-annotated theorem proof in training split, we sample $w=8$ tactics 
% $\{a_t^j \sim \pi_{ref}(a|s_t, p_t)\}_j^w$
$\{a_t^j \sim \pi_{ref}(a|s_t)\}_j^w$ 
% \todo{I removed the premises $p_t$ from the conditional distribution to simplify}
via \emph{beam search} and sort them by their likelihood score given by the sampling model. Additionally, we look for the presence of the ground truth tactic, i.e., gold tactic, in the beam search proposals: if the golden tactic 
$\widehat{a_t}$ for the current state $s_t$ happens to be absent from the top $w$ proposals, we append it as a $w+1=9$-th tactic at the end of the list (i.e. with lowest likelihood); this means states can have either $8$ or $9$ tactic proposals for training. The next step involves accruing feedback from the Lean framework by individual application of each proposed tactic on the current proof state.

\paragraph{Dataset for DPO}
To construct the dataset of \emph{pairwise} comparisons $D$ required by DPO, for each state $s_t$, we need to label its tactics as the positive and negative samples using Lean feedback. A tactic is positive if it is syntactically correct and applicable to the current state $s_t$ by Lean, and it is negative if it gets rejected by Lean.
We consider three strategies in the pair creation.
\textbf{random}: We pair each negative tactic with a positive tactic randomly. % we pick a we randomly pick positive and negative tactics for the given state.
\textbf{zero accuracy}: Each negative tactic is paired with the highest-likelihood positive one among those mistakenly ranked lower.
\textbf{hard}: Similar to zero accuracy, but we pick the lowest-likelihood positive tactic.
In practice, we also include some minor heuristics to avoid resampling the same positive tactics too often, see Algorithm~\ref{alg:dpo_dataset}.
Finally, for data augmentation purposes, rather than using the original prompt comprising the state and the retrieved premises $x=(s_t, p_t)$, we dynamically generate a new prompt like in \cite{yang2023leandojo} by applying random dropout on the retrieved premises.
Then, once a positive tactic $y^+$ is chosen, and the augmented prompt $x$ is generated, we add the tuple $(x, y^+, y^-)$ to the dataset $D$.
Each proof state $s_t$ appears in the dataset for each negative tactic the reference model generated for it. 
In the end, the offline preference dataset consists of $251\text{k}$ triplets $(x, y^+, y^-)$. For online DPO, we keep the same procedure, except for the reference model used for generating the beam-search tactic proposals per proof step, which gets updated every $50$ iterations with the target model. We describe precisely this dataset generation procedure in Algorithm~\ref{alg:dpo_dataset}.

\looseness=-1
\paragraph{Dataset for GRPO}
For the GRPO scenario, the dataset is being built dynamically throughout the training. More specifically, for each state in the training set, the model generates a set of tactics using beam search. Then, using the reward function described in Sec. \ref{sec:methodology}, each generated tactic is scored. The more (sub)goals a tactic solves, the greater its reward. Since the policy model is getting updated through GRPO objective, the model's generated tactics change over time. Similar, to the online case in DPO, we update the model used for sampling tactics every $50$ training iterations with the on-policy target model.

\subsubsection{Training Setting}
We use the pre-trained generator model in ReProver from \cite{yang2023leandojo} both as a base reference model for $\pi_{ref}$ and to initialize the policy to optimize $\pi_\theta$. The generator is an encoder-decoder Transformer based on the ByT5 model \cite{xue-etal-2022-byt5}, accompanied by a premises retrieval system that provides the most relevant premises as input. We fine-tune the pre-trained ReProver model using the described datasets for $10\text{k}$ steps with the AdamW optimizer (learning rate $2.25e-6$, batch size 16). The overall fine-tuning process in the LeanListener framework requires approximately 40 A100 GPU-days, with over 80\% of the time spent on interactions with Lean, which are CPU-bound. All experiments were conducted using HuggingFace TRL trainers.

\begin{table*}[th]
    \centering
    \small
    \begin{tabular}{llcc|cc}
         \textbf{Model}             & \makecell{Policy\\ Opt. Method} & \makecell{Pairing \\ Strategy} & Online   & \texttt{random} & \makecell{\texttt{novel} \\\texttt{premises}} \\ %\textbf{Pass@1} ($\%$) \\
         \hline
         \texttt{tidy }                                                                     &&&& $23.8$          &  $5.3$               \\
         GPT-4                                                                              &&&& $29.0$          &  $7.4$               \\ 
         ReProver (w/o retrieval)                                                           &&&& $47.6$          &  $23.2$               \\
         ReProver                                                                           &&&& $51.2$          &  $26.3$               \\
         ReProver*                                                                          &&&& $52.76$         &  $40.86$                \\
         \hline                          
         \multirow{7}{*}{LeanListener (Ours)}        & \multirow{6}{*}{DPO (binary)}                                
                                                                 & rand.            & $\times$ & $35.99$         &                   \\
                                    &                            & zero acc.        & $\times$ & $33.18$         &                   \\
                                    &                            & hard             & $\times$ & $31.27$         &                   \\
         \cline{3-6}                                        
                                    &                            & rand.            & $\vee$   & $50.25$         &                 \\
                                    &                            & zero acc.        & $\vee$   & $50.90$         &                 \\
                                    &                            & hard             & $\vee$   & $50.85$         &                 \\
         \cline{2-6}              
                                    & GRPO (\#sub-goals)         & -                & $\vee$   & $\boldsymbol{53.21}$   & $\boldsymbol{41.11}$            \\
         \hline
    \end{tabular}
    \caption{
        \textbf{Pass@1} ($\%$) performance on the LeanDojo benchmark on the \texttt{random} and \texttt{novel premises} splits. The performance of the first four baselines is the one reported in \cite{yang2023leandojo}, while ReProver* is the newly provided pre-trained model, which we evaluated ourselves.
    }
    \label{tab:leandojo}
\end{table*}

\begin{figure*}
    \centering
    \scalebox{0.9}{
    \subfigure[\texttt{random} split]{
        \includegraphics[width=0.49\linewidth]{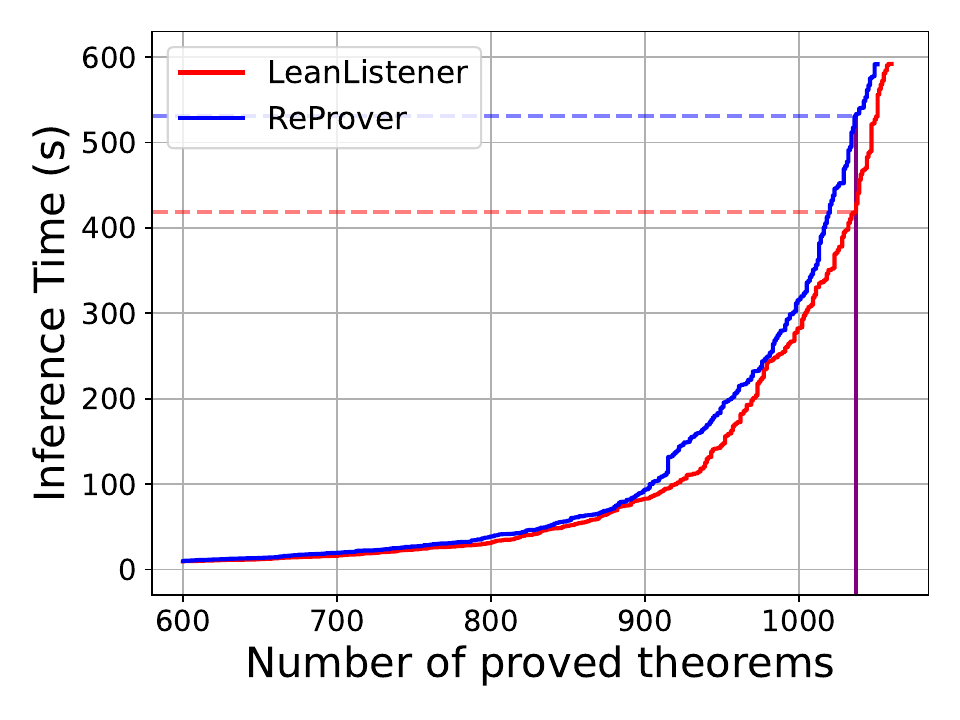}
        \label{fig:inference_time1}
    }%
    % \hspace{0.1\textwidth}
    \subfigure[\texttt{novel premises} split]{
        \includegraphics[width=0.49\linewidth]{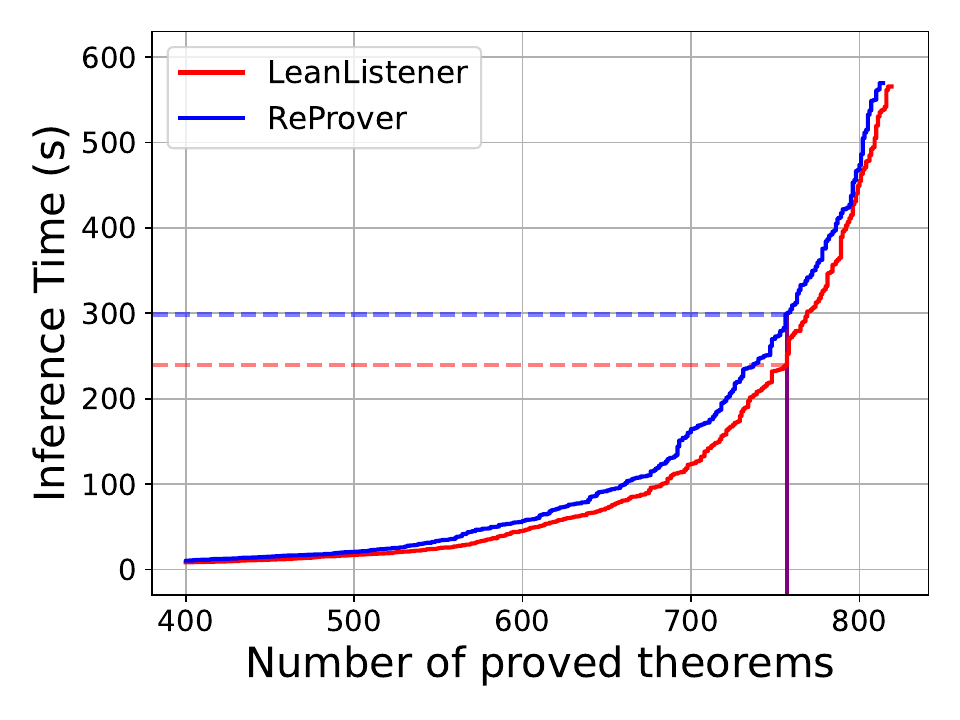} 
        \label{fig:inference_time2}
    }}
    \caption{To understand the improvement in inference speed, in these figures, we visualize the number of theorems in the two test sets which are proven by each model within different inference time limits. While the overall number of theorems proved within $10$ minutes by the two models is comparable, we see \textit{LeanListener} is consistently faster, which means, it takes significantly less time to solve the same number of theorems (see the purple lines for example). }
    \label{fig:inference_time}
\end{figure*}
\section{Experiments}
\label{sec:experiments}

\looseness=-1
We evaluate the effectiveness of LeanListener in different aspects on the test set of LeanDojo benchmarks. The test set has two variants, \texttt{random}, where the whole dataset is randomly split into training, validation, and test set, and the more challenging one, \texttt{novel premises}, where the test set includes premises that do not appear in the training data\footnote{
\cite{yang2023leandojo} only released the model trained on the \texttt{random} split, which we use as the base for our method.
We report the performance of the methods also on the \texttt{novel premises} test despite the possible overlap with the training set since this testset is still more challenging than the \texttt{random} one.
See \href{https://github.com/lean-dojo/ReProver/discussions/51\#discussioncomment-10767884}{github issue}.}. In the following experiments, we first verify whether incorporating Lean feedback enhances the quality of generated tactics. Then, we examine the performance on theorem proving task. Finally, we study the improved theorem proving speed of our method.

\looseness=-1
\paragraph{Tactic validity.}
To begin with, we investigate to what extent adding DPO and GRPO in the training pipeline increases the number of valid tactics generated by the model. As mentioned before, a tactic is valid if it is applicable to a state by Lean, even if it does not lead to a proof. The validity is also about the syntactic correctness of the tactic. We expect that if the model generates more valid tactics in the best first search, the chance of proving the given theorem, as well as the efficiency of the proof, will increase.  

Table \ref{tab:stepwise_stats} presents the numerical results on the validity of the generated tactics under different scenarios using DPO and GRPO objectives.
In this table, we compare the state-wise metrics which we expect to correlate with our training objective.
First of all, using the DPO objective, we observe that all offline strategies fail to improve the number of valid tactics generated by the model, while the online strategy significantly improves the number of valid tactics, and reduces the number of states with no valid tactics. Among all, online GRPO provides the best results. It is interesting to observe that the length of valid tactics decreases in the best models, thereby reducing unnecessarily complex tactic generation. 
% Additionally, we note that our training strategy significantly increases the length of the generated tactics, indicating that the model generates more complex tactics.
% {\color{red}Explaining GRPO results......}

% In our experiments, we compare our method with the ReProver model on the original LeanDojo benchmark \cite{yang2023leandojo}.

% Previously, we introduced the DPO objective in order to increase the number of valid tactics generate by the model.
% Hence, we first investigate the effectiveness on this task.
% Table~\ref{tab:stepwise_stats} compares the ReProver baseline \cite{yang2023leandojo} before and after tuning with different variations of our DPO method.

% \subsection{Performance evaluation}
\paragraph{Performance evaluation.}  Moving forward, we investigate whether improved tactic generation results in stronger reasoning capabilities or not.
Table~\ref{tab:leandojo} presents the final theorem proving the performance of our models with the pre-trained ReProver baseline along with GPT-4 \cite{Achiam2023GPT4TR} and \texttt{tidy} (which is a non-machine learning and heuristic-based approach), on the LeanDojo benchmark when using the \emph{random} split. \footnote{The evaluation results on the MiniF2F \cite{zhengminif2f} and ProofNet \cite{azerbayev2023proofnet} datasets are reported in Appendix \ref{app:other-results}.}.
We first observe that while DPO training on binary feedback improves the number of valid tactics, it always leads to a small degradation in the number of proven theorems. 
We believe that there is a spurious correlation between the low-ranked valid tactics and their length in the paired samples in DPO training, which leads to a bias toward generating longer and not necessarily useful tactics.
% this effect is due to the fact that lower-ranked tactics are typically longer: by simply preferring lower-ranked valid tactics, the model mistakenly also learns to generate longer tactics, which can be more prone to failure.
GRPO effectively addresses this issue by online training and avoiding paired samples for training, as well as employing more informative feedback about the number of solved sub-goals.
Indeed, we find that this finer reward, which incorporates the number of solved sub-goals, is more effective and leads to an increased number of proven theorems.

\begin{figure*}[h]
    \centering
    \subfigure[Per-theorem proof-length analysis]{
        \includegraphics[width=0.49\linewidth]{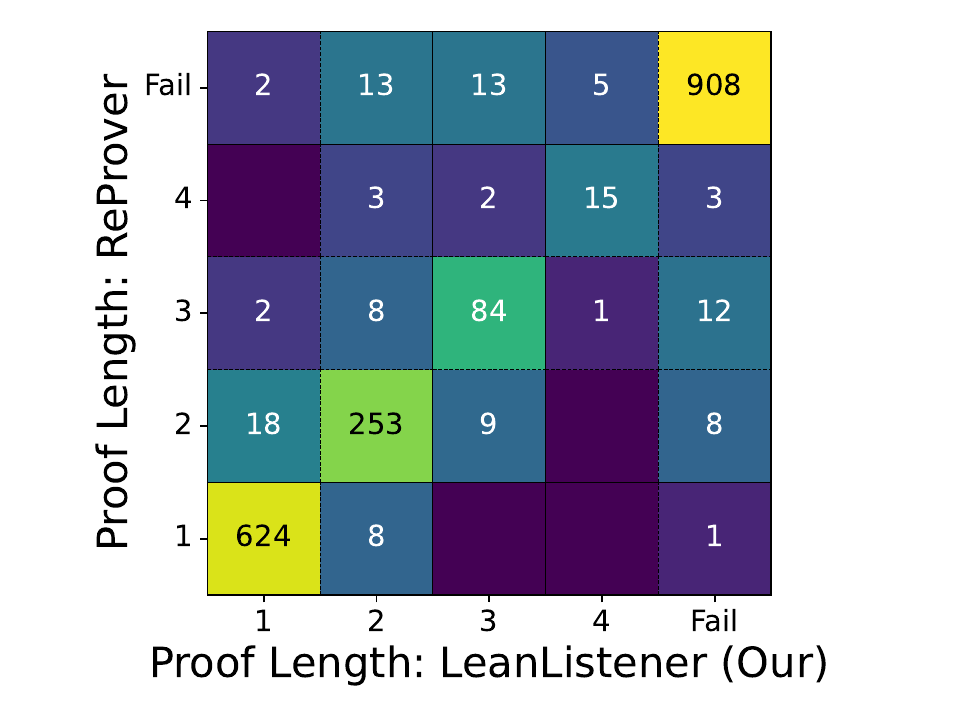}
        \label{fig:proof_len_adv}
    }%
    % \hspace{0.1\textwidth}
    \subfigure[Per-theorem relative proof-length comparison]{
        \includegraphics[width=0.49\linewidth]{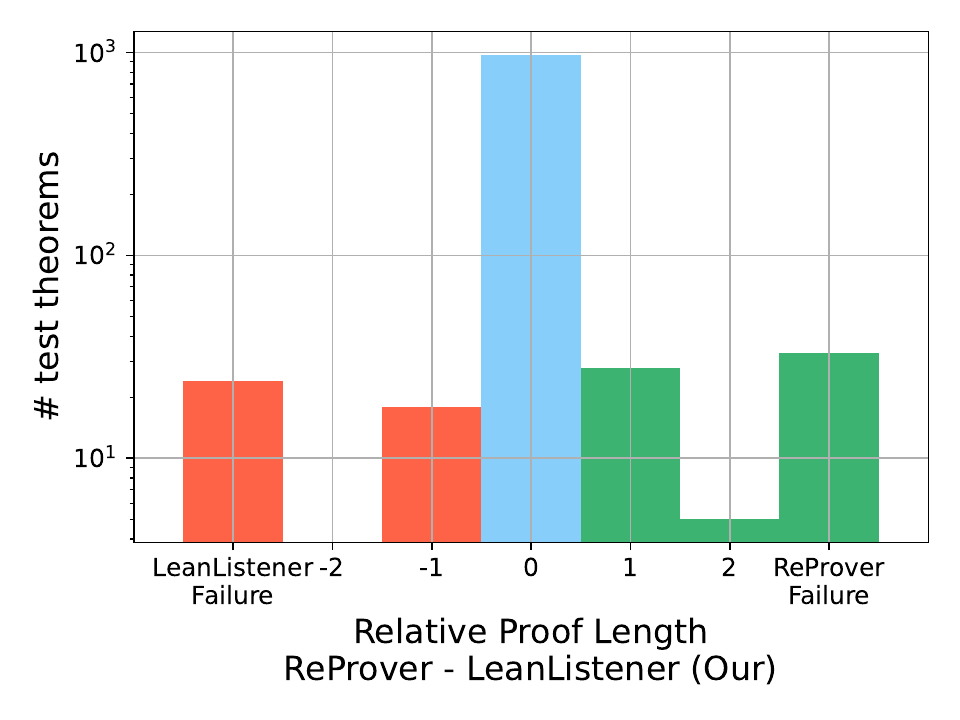} 
        \label{fig:proof_len_rel_adv}
    }
    \caption{For each valid theorem in the LeanDojo test set, we plot and compare the proof-length performance for both the \textit{LeanListener} and \textit{ReProver} models. \ref{fig:proof_len_adv} depicts the proof lengths for the proofs generated by the \textit{ReProver} model on the $y$-axis, and by the \textit{LeanListener} model on the $x$-axis. The plot shows a higher density above the diagonal, indicating that the proofs generated by \textit{LeanListener} generally have fewer proof steps than those generated by \textit{ReProver} for the same theorem. \ref{fig:proof_len_rel_adv} illustrates the relative proof-length advantage of the \textit{LeanListener} method over \textit{ReProver}. For each theorem, we plot the difference in proof lengths generated by both models, observing that instances where \textit{LeanListener} has a positive advantage outnumber those where it does not.}
    \label{fig:proof_len}
\end{figure*}

\paragraph{Inference time.}
In the previous experiments, we show that LeanListener generates a more efficient search tree by having more valid tactics. Now, we want to evaluate if this approach leads to a more efficient inference time. Figure \ref{fig:inference_time} compares the inference time in LeanListener and ReProver as the baseline model on the LeanDojo benchmark. As can be observed, LeanListener proves more theorems in up to $20\%$ less time for both random and novel premises splits. 

\paragraph{Proof length}
 In addition to the aggregate Pass@1 performance metric discussed earlier, we present an analysis of the proof lengths generated by the LeanListener and ReProver models on the LeanDojo benchmark\footnote{See Appendix \ref{app:other-results} for the analysis on MiniF2F and ProofNet.}. Generally, concise proofs, i.e., those with fewer proof steps, are considered more efficient and desirable than lengthier ones. Our aim here is to provide insights into the quality of proofs generated by both models. As shown in Figure \ref{fig:proof_len}, our LeanListener consistently produces more concise proofs compared to the base ReProver model.

An interesting observation in Figure \ref{fig:proof_len_adv} is the last row, which corresponds to proofs that were proved in just one step by the ReProver model. In this category, LeanListener fails to prove only one theorem, which is intuitive. Since LeanListener is trained with a one-step look-ahead process, it excels at proving theorems that can be solved in a single step. We observe that the number of failed proofs increases for theorems requiring more than one proof step. This suggests potential for further improvements in LeanListener's performance with training that incorporates more than one step look-ahead.

\section{Conclusions}
\looseness=-1
In this paper, we propose a novel RL-based framework, \emph{LeanListener}, based on the provided feedback by an external automatic verifier to improve trained LLMs in automatic theorem proving. LeanListener considers the general applicability of tactics as well as their local effectiveness, i.e., their impact on the number of unproven (sub)goals in a proof sequence, to fine-tune theorem provers.
Such step-wise reinforcement with the corresponding reward can efficiently harness the benefits of the trajectory level guidance without outcome dependent reward computation. %with efficient inference time computation. 
Our experimental results show that LeanListener not only surpasses the considered baselines in proving more theorems on the LeanDojo benchmark but also does so faster in less inference time.  

\newpage
\section{Limitations}

In this work, we have focused on a well-known open-source ITP model to show the effectiveness of the LeanListener framework. Applying our framework to more tactic generator models with different architecture, model sizes, and components in the training pipelines is an interesting future direction. We also mainly study the performance of LeanListener on the LeanDojo benchmark. Investigating the performance of LeanListener on other theorem proving datasets is another potential follow-up work.

\bibliography{bibliography}

% \printbibliography

%%%%%%%%%%%%%%%%%%%%%%%%%%%%%%%%%%%%%%%%%%%%%%%%%%%%%%%%%%%%%%%%%%%%%%%%%%%%%%%
%%%%%%%%%%%%%%%%%%%%%%%%%%%%%%%%%%%%%%%%%%%%%%%%%%%%%%%%%%%%%%%%%%%%%%%%%%%%%%%
% APPENDIX
%%%%%%%%%%%%%%%%%%%%%%%%%%%%%%%%%%%%%%%%%%%%%%%%%%%%%%%%%%%%%%%%%%%%%%%%%%%%%%%
%%%%%%%%%%%%%%%%%%%%%%%%%%%%%%%%%%%%%%%%%%%%%%%%%%%%%%%%%%%%%%%%%%%%%%%%%%%%%%%
\newpage
\appendix

\onecolumn

\section{Extended Discussions on Related Works}
\label{supp:related_works}

\begin{table}[ht]
    \centering
    \begin{tabular}{|c||c|c|c|c|c|}
       \hline
        Reference & Verifier & Task & Output & \makecell{Premise\\ Retrieval} & \makecell{Proof\\ Search}\\
       \hline
       \hline
       \makecell{GPT-f \\ \cite{Polu2020-gptf} } & Metamath & Prover & Proof Step & - & \makecell{Best First \\ Search}\\
        \hline
        Lean-Gym
        \cite{poluformal} & Lean & Prover & Proof Step & - & \makecell{Best First \\ Search\\ Expert \\ iteration}\\
        \hline
       \makecell{LeanDojo \\ \cite{yang2023leandojo}} & Lean & Prover & Proof step & \makecell{RAG \\ Offline \\ Database}& \makecell{Best First \\ Search} \\
       \hline
       \makecell{LeanAgent \\ \cite{kumarappan2024leanagent}} & Lean & Prover & Proof step & \makecell{RAG \\ Online \\ Database} & \makecell{Best First \\ Search} \\
        \hline       
        \makecell{\ac{HTPS} \\ \cite{lample2022hypertree}} & \makecell{Lean\\Metamath} & Prover & Proof step & - & HTPS \\
        \hline
        \makecell{DeepSeek-Prover \\ \cite{xin2024deepseek} } & Lean &  \makecell{Prover\\ Reasoning} & Whole Proof & - &  - \\
        \hline
        \makecell{DeepSeek-Prover 1.5 \\ \cite{Xin2024-deepseekproverv15} } & Lean &  \makecell{Prover\\ Reasoning} & \makecell{Proof Step \\ Whole Proof} & - & MCTS \\
        \hline
        % \makecell{DeepSeekMath \\ \cite{Shao2024-deepseekmath} }& - & Reasoning & Text & & \\
        % \hline
        \makecell{LLEMA \\ \cite{azerbayev2024llemma}} & Any Tool & \makecell{Prover\\ Reasoning} & \makecell{Proof \\ Text} & - & \makecell{Best First \\ Search}\\
        \hline        
        \makecell{Verify Step by Step \\ \cite{Lightman2023-s2sverify}} & \acs{ORM}-\acs{PRM} & Reasoning & Text & - & \makecell{Best First \\ Search}\\
        \hline
    \end{tabular}
    \caption{ A summary of models for theorem proving and mathematical reasoning
}
    \label{tab:comparison_provers}
\end{table}

\paragraph{Theorem Proving.} Machine learning is currently used in automatic theorem proving to generate 
the whole proof, or provide assistance in  sequential  proofs by finding the right premise or right tactics, or assisting in theorem prover specific formalization of mathematical statements, as well as combination of these ideas; see  \cite{xin2024deepseek,Xin2024-deepseekproverv15, Polu2020-gptf, Jiang2022-Thor, Wang2023-LEGO,Jiang2022-DraftSketch,Wu2022-Autoformalization} for some pointers. We review some of these works in more details. A key component for theorem proving is the formal proof management systems and verifiers like Lean \cite{Moura2021-Lean4} and Coq \cite{Coq2024}. Lean has received particular attention in the mathematics community for example by its use in verification of some components in the condensed mathematics research program \cite{Scholze2022-qj}. We use Lean in this work, and a framework that consists of a premise retrieval part and a prover that generates step-wise proof tactics.

Machine learning can be used to find useful premises, definitions and theorems \cite{Irving2016-DeepMath}. 
For Coq framework \cite{Coq2024}, the authors in \cite{Blaauwbroek2024-km} introduce a graph neural network for embedding the new definitions using a hierarchical representation based on graph representations of theorems and definitions in  the Tactician platform \cite{Blaauwbroek2024-xa}. This enables them to use recent proofs and theorems in Coq, while kNN is used for the more recent tactics written by the user. In Thor \cite{Jiang2022-Thor}, the authors introduce a class of methods to use automated theorem provers for premise selection. LeanDojo \cite{yang2023leandojo} provided a framework that retrieves the premise from a database of Lean premises and uses ReProver to generate tactics for theorem proving in Lean. 
LeanAgent \cite{kumarappan2024leanagent} adds a dynamic database to LeanDojo, which enables the continual learning of the agent. Besides, a new curriculum learning based on the difficulty of the theorems is also added. 

Searching over different proofs and tactics is another component explored in various works, for example \cite{poluformal,Xin2024-deepseekproverv15,lample2022hypertree}. As discussed in \cite{poluformal}, any \ac{RL} algorithm for theorem proving should address two challenges, namely an infinite dimensional action space, and the absence of a natural opponent for self-play. Therefore, the authors suggest using expert iteration \cite{anthony2017thinking}, which amounts to iteratively fine-tuning a based model and searching to generate correct proofs. They also introduce \texttt{lean-gym} to facilitate the search procedure in Lean.  \acp{HTPS} was introduced in \cite{lample2022hypertree} which is a new search algorithms within an online reinforcement learning procedure. The key idea is that the search is represented over a hypergraph, where a policy network generates tactics composed of sub-goals, and then each goal is expanded for proof using new tactics. The provability of each goal is approximated using a critic.  The idea of keeping the visit counts, action values and their statistics follows the similar MCTS procedure. In \cite{gloeckle2024abel}, the authors introduce a reinforcement learning based framework for theorem proving in Lean 4, which consist of using  a programming interface based on Aesop for proof search organization, the \ac{HTPS} (\cite{lample2022hypertree}) procedure with an online reinforcement learning step.  In \cite{Zhao2023-enigma}, the authors use subgoal learning from reinforcement learning to decompose an informal LLM generated proof into subgoals and verify using a verifier. They also use a diffusion model for demonstration organization. 

The transformer based language models are used also for theorem proving. For example  GPT-\emph{f} was introduced in \cite{Polu2020-gptf} using Metamath as the formalization framework. Besides, they showed that iterative training a value functionon proofs generated by the prover can continually improve the performance. Another example is LLEMMA which is based on pretrained Llama Code \cite{azerbayev2024llemma}. 

The proofs in some frameworks are already in the language of formal theorem provers. 
Besides, the external tools for mathematical reasoning can include codes for mathematical arguments. 
Machine learning has been used to help the formalization of mathematical statements suitable for the automatic theorem provers \cite{Wu2022-Autoformalization,Jiang2022-DraftSketch,Zhao2023-enigma}. Furthermore, there is a spectrum of methods that starts here with methods operating in a formal language and ends with \textit{informal} proofs and solutions in natural language. In MathCoder2 \cite{Lu2024-vx}, the authors provided pairs of mathematical codes and the associated natural language versions. 

% \todo{I have moved this paragraph to the main paper, since DeepSeekProverv1.5 uses GRPO as RL algorithm to train whole proof using verifier feedback, so it is very related to our method}
DeepSeek-Prover \cite{xin2024deepseek} generated Lean4 proof data by translating natural language problems into their formal versions in Lean4. The model produces the whole proof in a single turn. DeepSeekV1.5 \cite{Xin2024-deepseekproverv15} provides a middle ground by generating the whole proof, verifying it via Lean, and then truncating the proof until the first error. The authors propose additional techniques relying on proper appending of previous states and including truncate-and-resume in the \ac{MCTS} procedure.
They also leverage Verification feedback from  Lean and  improve model's alignment with the formal structure of Lean. The authors in \cite{Wang2023-LEGO} provide another hybrid solution where first an informal proof is generated, then broken into sub-components, and then a lemma is retrieved from a library of lemmas considered as skills, and finally the final formalized proof is generated using the previous components. They use the framework of Isabelle \cite{Paulson1994-Isabelle} for theorem proving.

Treating mathematical problem solving as a reasoning task has been considered in many works, for example \cite{Tong2024-DART-math,Zhu2023-cooperative-reasoning,Shao2022-cot-numerical, Shao2024-deepseekmath,Guan2025-rstar-math}. Among them, there are models, like \cite{Shao2024-deepseekmath}, based on chain-of-thought \cite{Wei2022-cot-prompting} or program of thought \cite{Gao2023-pal,Chen2022-pot}. The generator-verifier configuration has been considered in \cite{Zhu2023-cooperative-reasoning}, where they train a step and path verifier for reasoning. 
Integrating a tool for improving reasoning of large language models has been used in \cite{Gou2023-TORA-tool}. In contrast to our work, these models do not rely on theorem provers.  The authors in 
\cite{wang2020learning} address the lack of human labeled theorems and proofs to use for supervised training by training a generative model and, then, use the synthesized pairs to train a theorem prover model.

The idea of proof size and using it for guiding the search has been discussed in \cite{Wu2021-proofcost,poluformal}. There is a line of work related to discovering new mathematical functions or solving mathematical problems, see for example \cite{Real2023-autonumerics,Alfarano2024-lyapunov,Marchetti2023-neural-lattice,Wenger2022-lattice-crypto}. However, these works do not rely on proof verifiers and fall outside the scope of this work.

\paragraph{Preference Optimization.} Another important step in theorem proving, and many other alignment related tasks, is to \textit{align}  the output of a given model based on the positive-negative preference pair of samples. There are online version of \ac{RLHF} as in \cite{Bai2022-rlhf,Ouyang2022-rlhf}, and offline versions with either an explicit reward model (e.g. \cite{Christiano2017-human-preference,Ziegler2019-human-preference}) or an implicit reward model (e.g. \ac{DPO} \cite{rafailov2024direct})  to encode the preference. The later method directly optimizes the model without training any independent reward model. In particular \ac{DPO} uses Bradley-Terry preference model and the fact that the optimal solution to the KL-constrained reward maximization objective is known in closed form to simplify the training loss without dependence on a reward model \cite{rafailov2024direct}. Many follow-up works explored \ac{DPO} variations and its shortcomings such as reward over-optimization \cite{Gorbatovski2024-ref-model,Ethayarajh2024-kto}.

Different forms of direct preference optimization have been explored in the recent literature for enhancing the reasoning capabilities of language models, in particular for mathematical problem solving, for example \cite{Xiong2024-mulit-DPO,yuan2024advancing,jiao2024learning,Cobbe2021-train-verifier,Lightman2023-s2sverify,Wang2024-math-shepherd,pang2024iterative,chen2024step,lu2024step}. The authors in \cite{Xiong2024-mulit-DPO} introduce a multi-turn version of \ac{DPO} to use feedbacks from the verifiers, in their case the code interpreters, particularly for multi-turn scenarios, which requires trajectory-level preference optimization. The trajectory-level preference can be obtained by dataset labels of the gold answers or \acp{ORM} \cite{Cobbe2021-train-verifier,Lightman2023-s2sverify}. A more fine-grained step-wise supervision can be used as in \acp{PRM} \cite{Lightman2023-s2sverify} or  by leveraging trajectory level preferences \cite{Wang2024-math-shepherd}. The idea of \ac{ORM} and \ac{PRM} has also been originally discussed in \cite{Uesato2022-prm-orm}. These works focus on mathematical reasoning, rather then formal theorem proving, which is the focus of our work. In our work, the theorem prover automatically provides the preference, which is used either during training or for an offline generated dataset. Furthermore, our work follows \ac{PRM} philosophy as we consider feedback from Lean at each step of proof generation.

The authors in \cite{jiao2024learning, yuan2024advancing} applied the original \ac{DPO} or \ac{KTO} by taking trajectory completion as a meta action. The online iterative versions of  \ac{DPO} originally designed for chat is adapted to achieve better CoT reasoning in \cite{pang2024iterative,xie2024monte}.
In the papers \cite{chen2024step,lai2024step,xie2024monte,lu2024step}, the authors have explored generating proxy step-wise labels for the intermediate steps of the reasoning trajectories.

\section{Additional Details}

\begin{figure*}[t]
    \centering
    \scalebox{0.55}{
    \includegraphics{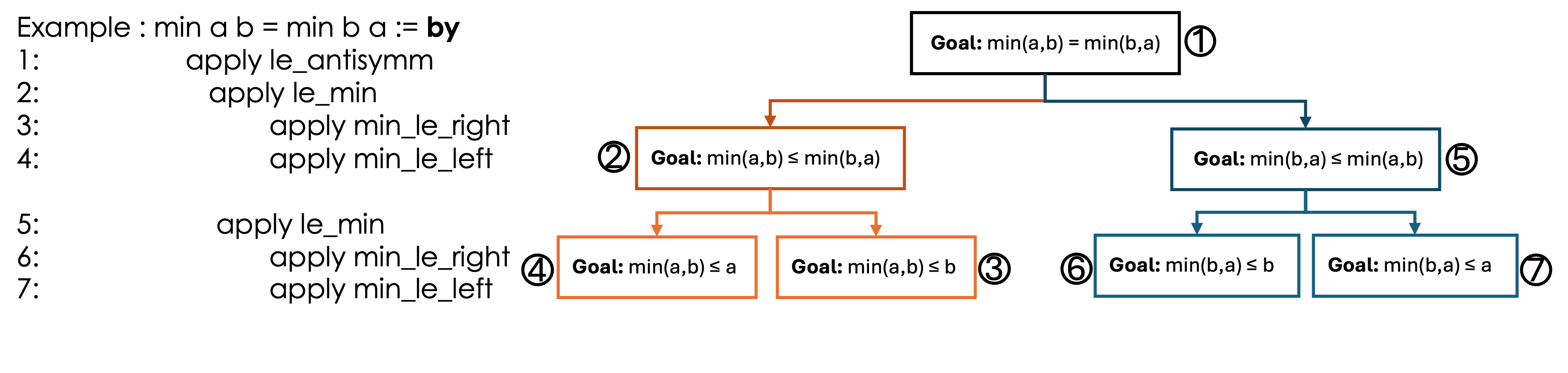}
    }
    \caption{A simple proof example in Lean on the left shows the $min$ function is symmetric, and its proof tree is at the right. The numbers next to the boxes refer to the corresponding line numbers in the written proof.}
    \label{fig:lean_example}
    
\end{figure*}

\begin{algorithm*}[th]
   \caption{DPO dataset generation using the \textbf{zero accuracy} strategy}
   \label{alg:dpo_dataset}
\begin{algorithmic}
   \STATE {\bfseries Input:} training dataset $D_\text{train}$, reference policy $\pi_\text{ref}$, retrieval model $\texttt{retriever}$
   \STATE $D \gets [ \ ]$
   \FOR{theorem and g.t. proof $(T, P) \in D_\text{train}$}
   \FOR{proof state and g.t. tactic $(s_t, \widehat{a_t}) \in P$}
        \STATE $p_t \gets \texttt{retriever}(s_t)$                                                            \COMMENT{compute the premises for retrieval-augmentation}
        \STATE $A_t \gets \{a^j_t \sim \pi_\text{ref}(a_t | s_t, p_t) \ |\ j=1, \dots, 8\}$          \COMMENT{with \emph{beam-search} (beam-width $w=8$)}
        \STATE sort($A_t$)                                                                                \COMMENT{by decreasing $\pi_\text{ref}(a | s_t, p_t)$ (if needed)}
        \IF{$\widehat{a_t} \notin A_t$} 
            \STATE append $\widehat{a_t}$ at the end of $A_t$ ($w=9$)  \COMMENT{add the ground truth tactic if needed}
        \ENDIF
       \STATE $A_t^+ \gets \{a_t^j \ |\ \text{Lean}(a_t^j | s_t, T) =\ + \}$                    \COMMENT{gather Lean feedback for each sampled tactic}
       \STATE $A_t^- \gets \{a_t^j \ |\ \text{Lean}(a_t^j | s_t, T) =\ - \}$                    
       \FOR{$y^- \in A_t^-$}
       \FOR{$y^+ \in A_t^+$}
            \IF{$\pi_\text{ref}(y^- | s_t, p_t) > \pi_\text{ref}(y^+ | s_t, p_t)$}
                \STATE $x \gets \text{dynamic\_prompt}(s_t, p_t)$                              \COMMENT{prompt augmentation by premises dropout}
                \STATE $D\text{.append(} (x, y^+, y^-)\text{)}$             
                \STATE move $y^+$ at the end of $A_t^+$ to lower its priority                   \COMMENT{simple heuristic to avoid oversampling $y^+$}
                \STATE {\bfseries break}  \COMMENT{exit the inner loop over $A_t^+$ and move to the next $y^-$}
            \ENDIF
       \ENDFOR
       \ENDFOR
   \ENDFOR
   \ENDFOR
   \STATE {\bfseries Output:}  $D$
\end{algorithmic}
\end{algorithm*}
\newpage
\section{MiniF2F and ProofNet Results}
\label{app:other-results}

\begin{table}[h!]
\centering
\begin{tabular}{l|cc}
\toprule
 & \textbf{MiniF2F} & \textbf{ProofNet} \\ 
 \midrule
ReProver & 26.5 & 13.8 \\ 
ReProver* & \textbf{37.7} & \textbf{15.3} \\ 
\midrule 
LeanListener & 36.9  & 14.4 \\ 
\bottomrule
\end{tabular}
\caption{\textbf{Pass@1} ($\%$) performance of the ReProver \cite{yang2023leandojo} and our framework, LeanListener, on MiniF2F and ProofNet. The ReProver* is the newly provided pre-trained model, which we evaluated ourselves.}
\label{tab:example}
\end{table}

\begin{figure*}[h]
    \centering
    \subfigure[Per-theorem proof-length analysis]{
        \includegraphics[width=0.49\linewidth]{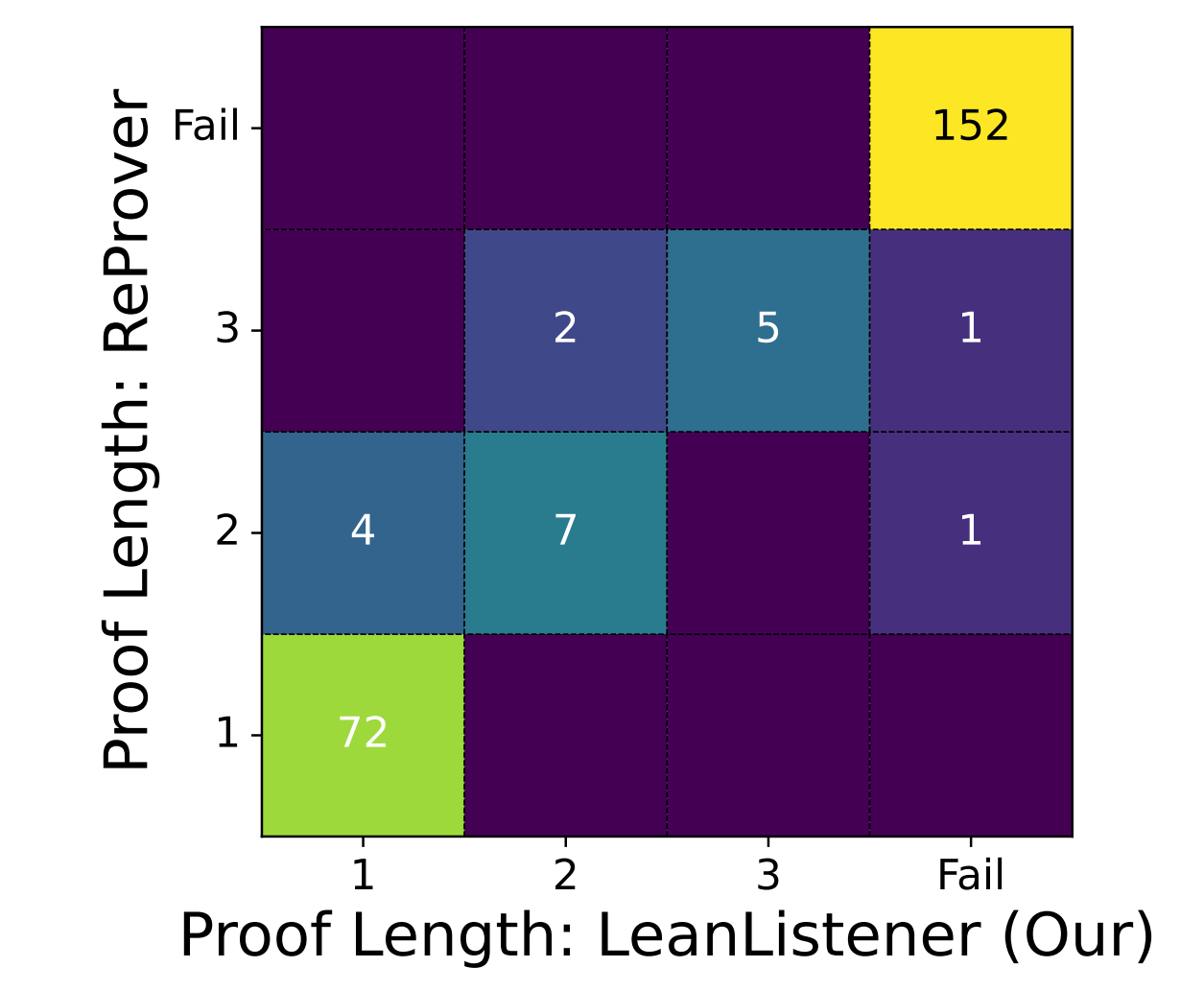}
        \label{fig:proof_len_adv_mini}
    }%
    % \hspace{0.1\textwidth}
    \subfigure[Per-theorem relative proof-length comparison]{
        \includegraphics[width=0.49\linewidth]{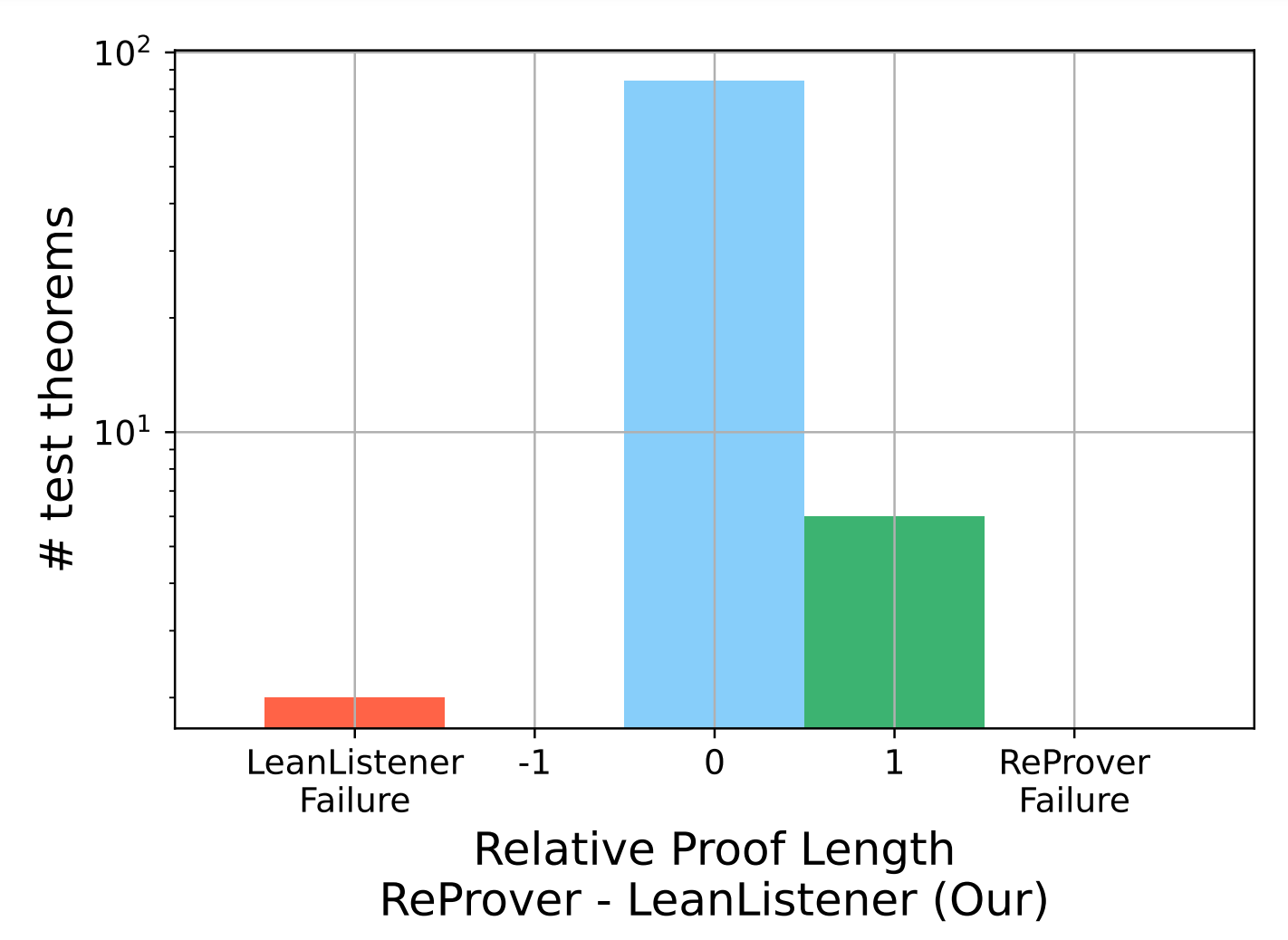} 
        \label{fig:proof_len_rel_adv_mini}
    }
    \caption{For each valid theorem in the MiniF2F test set, we plot and compare the proof-length performance for both the \textit{LeanListener} and \textit{ReProver} models. \ref{fig:proof_len_adv_mini} depicts the proof lengths for the proofs generated by the \textit{ReProver} model on the $y$-axis, and by the \textit{LeanListener} model on the $x$-axis. \ref{fig:proof_len_rel_adv_mini} illustrates the relative proof-length advantage of the \textit{LeanListener} method over \textit{ReProver}. For each theorem, we plot the difference in proof lengths generated by both models, observing that instances where \textit{LeanListener} has a positive advantage outnumber those where it does not.}
    \label{fig:proof_len_mini}
\end{figure*}

\begin{figure*}[h]
    \centering
    \subfigure[Per-theorem proof-length analysis]{
        \includegraphics[width=0.49\linewidth]{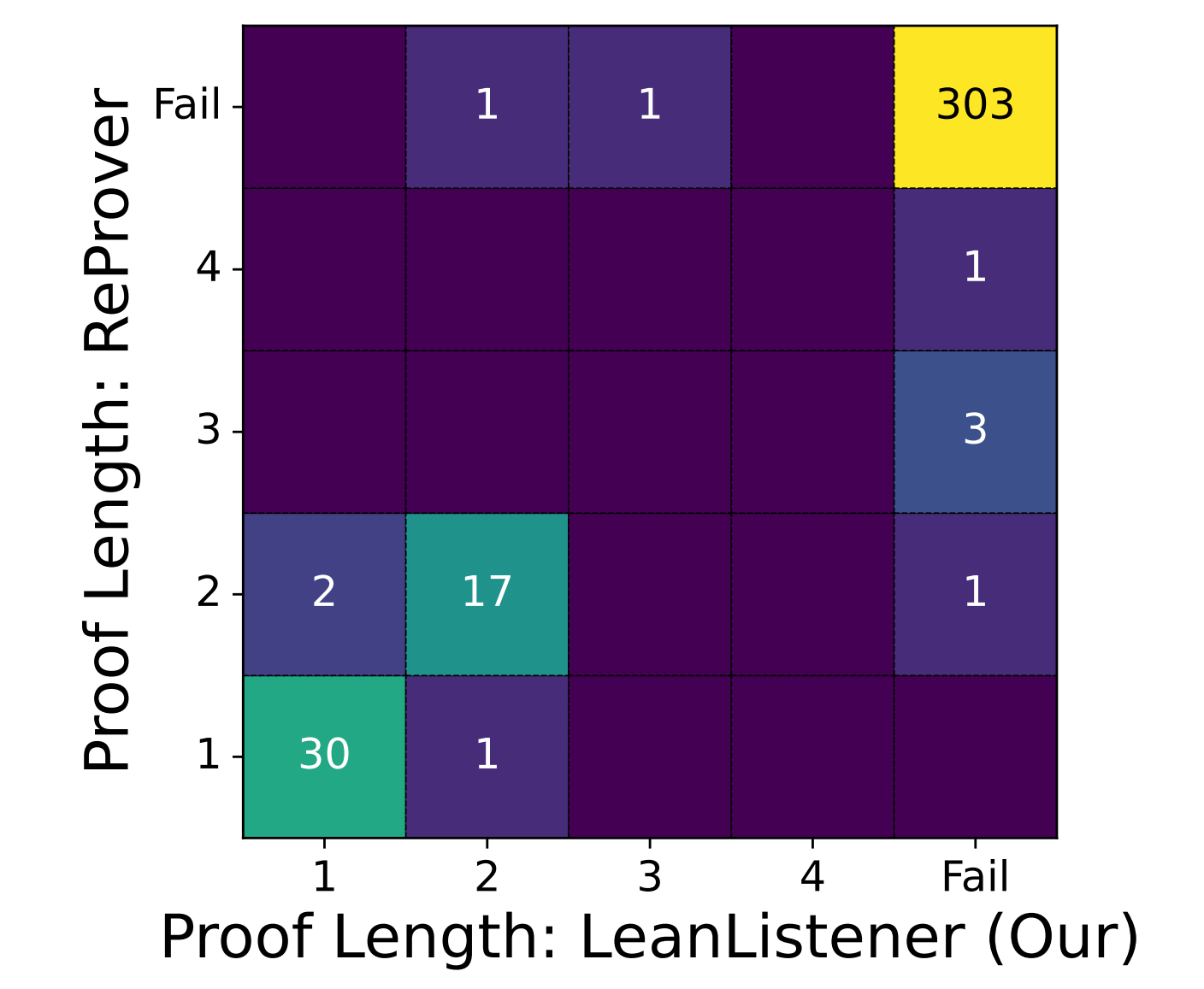}
        \label{fig:proof_len_adv_proofnet}
    }%
    % \hspace{0.1\textwidth}
    \subfigure[Per-theorem relative proof-length comparison]{
        \includegraphics[width=0.49\linewidth]{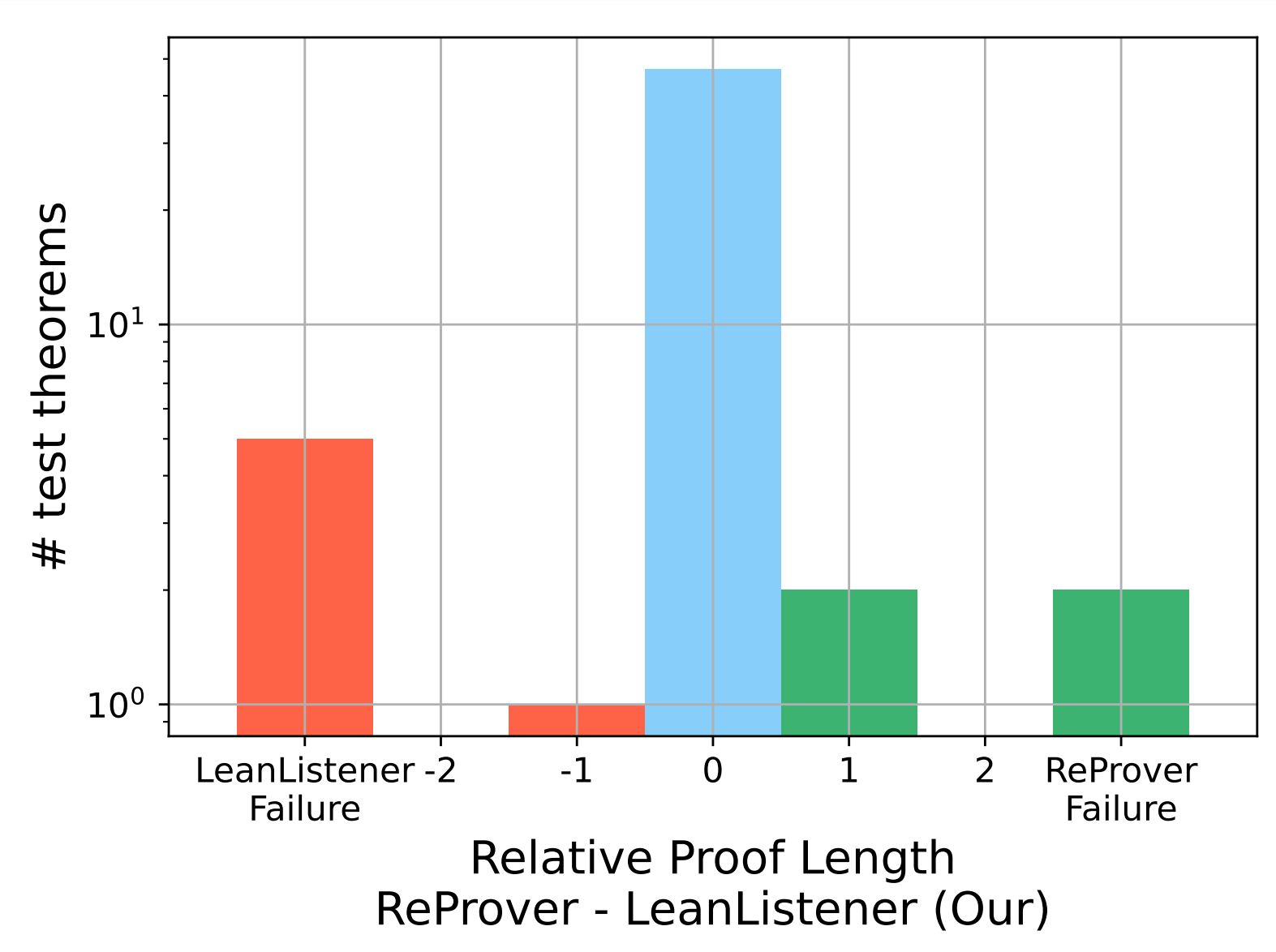} 
        \label{fig:proof_len_rel_adv_proofnet}
    }
    \caption{For each valid theorem in the ProofNet test set, we plot and compare the proof-length performance for both the \textit{LeanListener} and \textit{ReProver} models. \ref{fig:proof_len_adv_mini} depicts the proof lengths for the proofs generated by the \textit{ReProver} model on the $y$-axis, and by the \textit{LeanListener} model on the $x$-axis. \ref{fig:proof_len_rel_adv_mini} illustrates the relative proof-length advantage of the \textit{LeanListener} method over \textit{ReProver}. For each theorem, we plot the difference in proof lengths generated by both models, observing that instances where \textit{LeanListener} has a positive advantage outnumber those where it does not.}
    \label{fig:proof_len_proofnet}
\end{figure*}

%%%%%%%%%%%%%%%%%%%%%%%%%%%%%%%%%%%%%%%%%%%%%%%%%%%%%%%%%%%%%%%%%%%%%%%%%%%%%%%
%%%%%%%%%%%%%%%%%%%%%%%%%%%%%%%%%%%%%%%%%%%%%%%%%%%%%%%%%%%%%%%%%%%%%%%%%%%%%%%

\end{document}